\documentclass{article}

\usepackage[preprint]{neurips_2024}

\usepackage[utf8]{inputenc} 
\usepackage[T1]{fontenc}    
\usepackage{hyperref}       
\usepackage{url}            
\usepackage{booktabs}       
\usepackage{amsfonts}       
\usepackage{nicefrac}       
\usepackage{microtype}      
\usepackage{xcolor}         

\usepackage{microtype}
\usepackage{graphicx}
\usepackage{multicol}
\usepackage{subfigure}
\usepackage{booktabs} 
\usepackage{adjustbox}
\usepackage{pdflscape}

\usepackage{hyperref}
\usepackage{multirow}

\usepackage{amsmath}
\usepackage{amssymb}
\usepackage{mathtools}
\usepackage{amsthm}
\usepackage{booktabs} 
\usepackage{tabularx} 

\usepackage[capitalize,noabbrev]{cleveref}

\theoremstyle{plain}

\theoremstyle{definition}

\theoremstyle{remark}

\title{Serialized EHR make for good text representations}

\author{%
  Zhirong Chou\\
  Southern China University\\
  zc1222@scut.edu.cn
  \And
  Quan Qin\\
  Southern China University\\
  qq1782@@scut.edu.cn
  \AND
  Shi Li\\
  Columbia University\\
  shili081100@columbia.edu}
  
\begin{document}

\maketitle

\begin{abstract}
The emergence of foundation models in healthcare has opened new avenues for learning generalizable representations from large-scale clinical data. Yet, existing approaches often struggle to reconcile the tabular and event-based nature of Electronic Health Records (EHRs) with the sequential priors of natural language models. This structural mismatch limits their ability to capture longitudinal dependencies across patient encounters. We introduce \textbf{SerialBEHRT}, a domain-aligned foundation model that extends SciBERT through additional pretraining on structured EHR sequences. SerialBEHRT is designed to encode temporal and contextual relationships among clinical events, thereby producing richer patient representations. We evaluate its effectiveness on the task of antibiotic susceptibility prediction, a clinically meaningful problem in antibiotic stewardship. Through extensive benchmarking against state-of-the-art EHR representation strategies, we demonstrate that SerialBEHRT achieves superior and more consistent performance, highlighting the importance of temporal serialization in foundation model pretraining for healthcare.
\end{abstract}

\section{Introduction}

Foundation models represent a significant advancement in artificial intelligence, characterized by their ability to be used in a broad spectrum of tasks \citep{bommasani2021opportunities}. These models excel in representing complex information and creating latent representations via extensive pre-training, typically using self-supervised methodologies \citep{ericsson2022self}. Due to their robust data representation capabilities, foundation models have been applied to a wide range of tasks, such as content generation \citep{bandi2023power, radford2018improving, yang2023diffusion, cao2024survey}, natural language understanding \citep{devlin2018bert}, and applied across various fields in science and engineering \citep{lee2024can, lin2025case}.

In healthcare, these models are particularly promising, leveraging datasets for decision-making \citep{lee2025clinical, steinberg2023motor, wornow2023ehrshot}, summarization \citep{riccio2023healthcare, jin2024comprehensive}, and automatic coding \citep{soroush2024large}, among other utilities. Despite their versatility, foundation models face significant challenges due to the scarcity of clinical text. This underscores the urgent need for foundation models to represent granular tabular concepts \citep{van2024tabular}, such as patient events and encounter-level data often found in healthcare datasets.

To address these issues, this paper introduces \textit{SerialBEHRT}, building on the methodology of transforming Electronic Health Records (EHR) into text \citep{lee2025clinical} and using it for pre-training purposes. We leverage the established efficacy of foundation models trained on extensive textual data, providing a streamlined method for their pre-training \citep{kaplan2020scaling}. We validate our new foundation model, SerialBEHRT, by demonstrating its applicability on antibiotic stewardship, specifically in predicting suitable antibiotics for individual patients.

\section{Related Works}

\paragraph{Medical Representation Learning}

The field of medical representation learning has rapidly evolved, focusing on the analysis of Electronic Health Records (EHRs) to improve clinical decision-making processes. Initially, this process required substantial feature engineering to adapt raw EHR data for use with conventional machine learning models \citep{tang2020democratizing, ferrao2016preprocessing,wu2010prediction}. This method proved to be labor-intensive and exhibited considerable variation across different research teams due to the absence of standardized methods.

The current trend leans towards leveraging advanced foundation models that excel in representing medical data through the analysis of extensive textual datasets, including clinical notes, medical literature, and comprehensive patient records. These models are largely built upon the BERT architecture and are capable of producing deep, contextually aware latent representations of patient histories, which significantly diminish the reliance on manual feature engineering \citep{rasmy2021med,liu2021med,alsentzer2019publicly,lee2020biobert}. Additionally, recent advancements have introduced methods that integrate the temporal aspects of EHR data, using patient event/sequence histories to forecast medical outcomes more accurately \citep{steinberg2023motor, wornow2024ehrshot, pang2021cehr, li2022hi}.

\paragraph{Biomedical Language Models}

The field of medical representation learning has undergone significant advancements, with a focus on improving clinical decision-making by analyzing Electronic Health Records (EHRs). Initially, these efforts required extensive feature engineering to transform raw EHR data for use with traditional machine learning models \citep{tang2020democratizing, ferrao2016preprocessing, wu2010prediction}. However, this approach was time-consuming and lacked standardization, leading to variations in implementation across different research teams. In recent years, the trend has shifted towards leveraging advanced foundation models, particularly those based on the BERT architecture, which have proven highly effective in representing medical data by analyzing vast textual datasets such as clinical notes, medical literature, and patient records \citep{rasmy2021med, liu2021med, alsentzer2019publicly, lee2020biobert, lee2025clinical}. These models generate deep, contextually rich latent representations of patient histories, significantly reducing the need for manual feature engineering.

Given the success of the BERT model \citep{devlin-etal-2019-bert} in natural language processing, researchers have extended its utility to the biomedical domain, where records often exist in textual form. Early works, such as BEHRT \citep{li2020behrt} and MedBERT \citep{rasmy2021med}, demonstrated the potential of this approach. More recent models, such as ExBEHRT \citep{rupp2023exbehrt}, IRENE \citep{zhou2023transformer}, and M-BioBERTa \citep{antalm}, have continued to expand the variety of features and covariates available to the models, improving their contextual embedding processes. Additionally, models like TransformEHR \citep{yang2023transformehr}, Gatortron \citep{yang2022large}, and CLMBR \citep{wornow2023ehrshot} propose that using transformer decoders (with forward-directional attention) can enhance predictive accuracy in forecasting medical outcomes.

Transformer architectures have also been explored to create language models that reason with biomedical knowledge derived from scientific research \citep{NEURIPS2020_1457c0d6}. The BERT framework, in particular, has been applied to extract reliable representations of scientific concepts from research articles, as seen in models like SciBERT \citep{beltagy-etal-2019-scibert}, ClinicalBERT \citep{clinicalbert}, BioBERT \citep{lee2020biobert}, PubMedBERT \citep{gu2021domain}, and BioMegatron \citep{shin2020biomegatron}. These models are tailored to biomedical texts, further enhancing their performance on domain-specific tasks.

Recent developments in conversational language models, such as BioGPT \citep{luo2022biogpt}, highlight the potential of decoder-based architectures (forward-directional attention) for text generation tasks in the biomedical field. Studies, including \cite{kane2023compressed}, have shown that BioGPT provides state-of-the-art embeddings for diagnosis codes (ICD-10), improving predictions of their semantic relationships. These models have demonstrated a range of capabilities, from identifying clinical concepts in freehand medical text \citep{gu2021domain,ijcai2020-461-vu} to enabling interactive clinical dialogues \citep{varshney2023knowledge} and automating the writing of discharge notes \citep{ellershaw2024automated}. 

\paragraph{Text Serialization}

Text serialization was initially introduced by TabLLM \citep{hegselmann2023tabllm}, which transformed tabular datasets into text. Subsequent studies, such as those by  \citep{jaitly2023towards,dinh2022lift, ono2024text, an2025dk}, have explored various serialization strategies, data preprocessing techniques, and their comparative effectiveness against other machine learning approaches. These developments respond to the identified need for a foundational model capable of processing tabular data, as discussed in a recent review by Van et al. \citep{van2024tabular}. While there has been some progress in this area, exemplified by models like TaBERT \citep{yin2020tabert} and TabPFN \citep{hollmann2022tabpfn}, these models often face scalability limitations. Consequently, text serialization has emerged as a viable alternative for representing tabular data in textual form, a method that has been applied in the clinical context, as seen in the work of \citep{lee2025clinical}.

\section{Methods}

\begin{figure}
  \centering
  \includegraphics[width=0.8\textwidth]{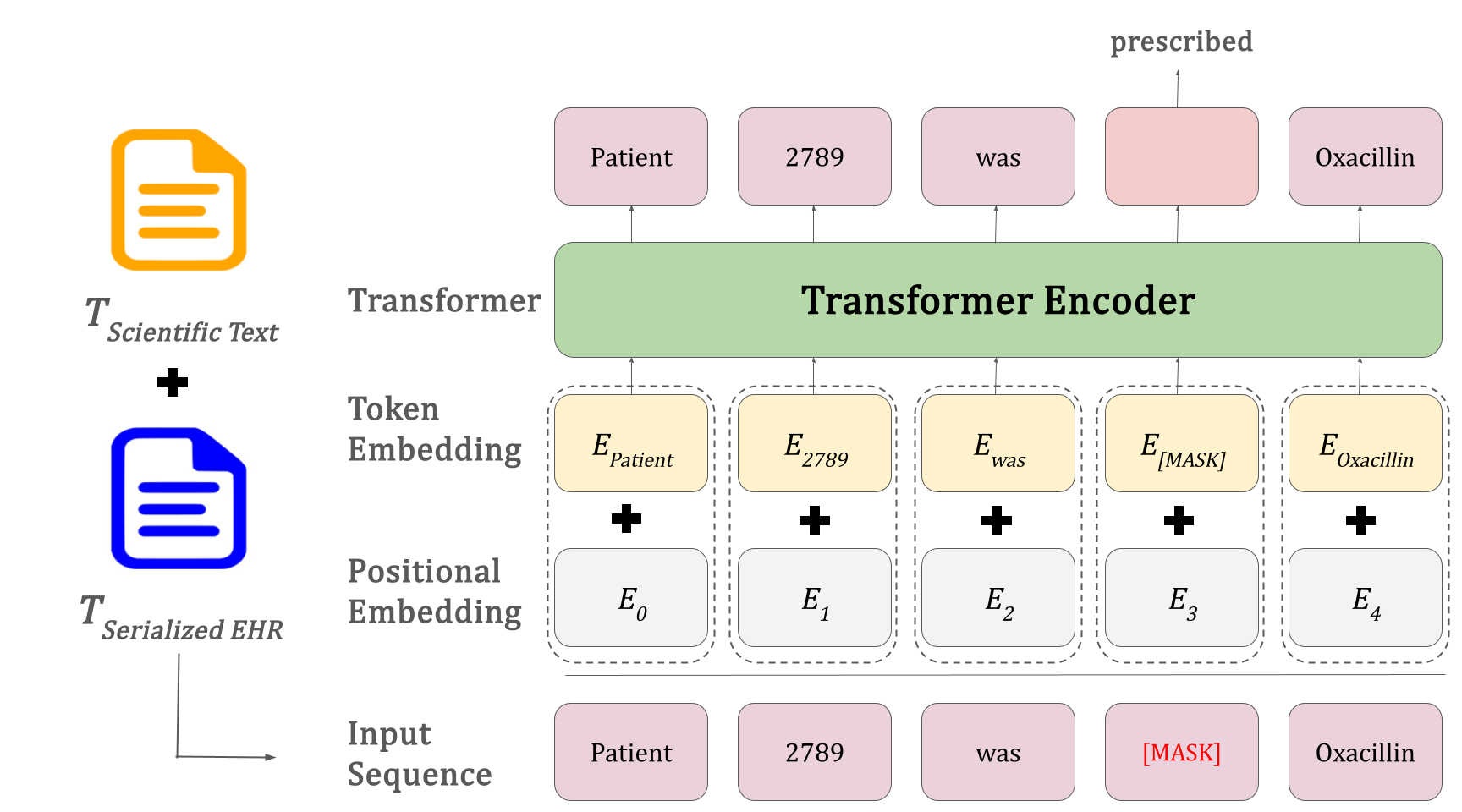}
  \caption{Schematic representation of SerialBEHRT, pre-trained on a composite of scientific texts and serialized EHRs, demonstrating the Masked Language Modeling (MLM) objective.}
\end{figure}

\subsection{SerialBEHRT}

SerialBEHRT marks an advancement in medical language modeling, building on the foundation established by SciBERT \citep{beltagy2019scibert}. While SciBERT was originally pre-trained on a corpus from Semantic Scholar \citep{ammar2018construction}, SerialBEHRT extends this by incorporating serialized Electronic Health Records (EHR) into its pre-training corpus. This integration is hypothesized to enhance the model’s ability to represent EHR concepts, as it exposes SerialBEHRT to clinical information that is often absent from standard clinical texts and discharge summaries.

The core contribution of SerialBEHRT lies in its ability to convert tabular EHR data \( X \) into a serialized textual format \( T \), as outlined in \citep{lee2025clinical}, which is then used for pre-training. This conversion facilitates training on a composite dataset that includes both scientific text \( T_{\text{Scientific\_Text}} \) and serialized EHR \( T_{\text{Serialized\_EHR}} \). The serialization process can be formally described as:

\[
X \rightarrow T(X) = \{ T(x_i) \mid x_i \in X \}
\]

In this formulation, \( X \) represents the original dataset, while \( T(X) \) indicates the transformation of \( X \) into serialized text, with each data entry \( x_i \) being converted into its text representation \( T(x_i) \). 

\subsubsection{Data Coverage}

SerialBEHRT leverages two distinct data sources: scientific text from Semantic Scholar \citep{ammar2018construction} and serialized electronic health records (EHR) \citep{johnson2023mimic}. The first comprises both structured and unstructured data from academic publications, which are rich in domain-specific knowledge and research findings. The second includes clinical data such as patient demographics, diagnostic codes, treatment information, and clinical outcomes, providing unique EHR-specific concepts. We hypothesize that these two types of text complement each other, thereby enhancing the breadth and depth of our model. Further details about these distinct concepts are presented in Table \ref{info-comparison-table}.

\begin{table}[h]
  \caption{Unique Elements of Scientific Texts (Journal Articles) vs. Electronic Health Records (EHRs)}
  \label{info-comparison-table}
  \centering
  \begin{adjustbox}{width=\columnwidth}
  \begin{tabular}{lll}
    \toprule
    \multicolumn{3}{c}{Information Type}                   \\
    \cmidrule(r){1-3}
    Description     & \textbf{Scientific Text ($T_{\text{Scientific Text}}$)} & \textbf{Electronic Health Records ($T_{\text{Serialized EHR}}$)}\\
    \midrule
    Objective & Aim to expand the scientific knowledge base & Focus on individual patient health and medical history \\
    Data Types & UnStructured around hypotheses, methods, results & Structured data from patient interactions \\
    Style & Formal, follows strict publishing guidelines & Practical, focused on clarity and immediate utility \\
    Updates & Published as new editions or separate articles & Continuously updated during patient care \\
    Length & Can be extensive, depending on the subject & Concise, limited to relevant medical details \\
    Purpose & Dissemination of new research findings & Supporting ongoing medical care \\
    \bottomrule
  \end{tabular}
  \end{adjustbox}
\end{table}

To quantify the overlap and distinctiveness between these datasets, we use the mathematical model \(\frac{|A \cap B|}{|A \cup B|} \ll 1\), where \(A\) and \(B\) represent the Semantic Scholar and serialized EHR datasets, respectively. This equation indicates minimal overlap, with the union of the datasets being significantly larger, suggesting that each dataset contains unique elements that are not present in the other. A formal description of both sources of data are described in Table \ref{info-comparison-table}. Additionally, an enhancement index \(\eta = \frac{|A \cup B|}{|A| + |B|}\), where \( \eta > 1 \) implies that the combined datasets provide more comprehensive coverage than the sum of the individual datasets. We model it as such because these two sources of information are hypothesized to be complementary and can provide context that is contained within each individual source. By integrating these diverse sources, we enhance the comprehensiveness of the pre-training corpora potentially leading to stronger representations for the clinical data.

\subsubsection{Vocabulary Expansion and Tokenizer Adaptation}
In order to account our new $T_{\text{Serialized\_EHR}}$ and its corresponding tokens, we introduce a methodology that improves SciBERT's tokenizer for the purposes of SerialBEHRT's. We accomplish this by expanding its vocabulary by training a SentencePiece byte-pair encoding (BPE) model \citep{kudo2018sentencepiece} on $T_{\text{Serialized\_EHR}}$. This training produced a domain-specific vocabulary $V_d$ of 10,000 tokens. We merged this with SciBERT's original vocabulary $V_o$ to create an expanded vocabulary $V_e = (V_o \cup \{t \mid t \in V_d, t \notin V_o\})$. This expansion not only retains SciBERT’s scientific terminology but also integrates new, EHR-specific tokens, thus encompassing both scientific and specialized medical terminologies.

Following this, we reinitialized a BertTokenizerFast \citep{song2020fast} with $V_e$, maintaining the original settings for case sensitivity and special tokens to preserve the pre-trained model’s tokenization behavior. The adapted tokenizer now efficiently processes domain-specific terminology from $T_{\text{Serialized\_EHR}}$ and expands the model’s vocabulary to $42,000$ unique tokens compared to its original $32,000$. This enhancement to the tokenizer ensures balanced recognition of both scientific and medical terms, effectively encoding clinical concepts unique to the EHR (e.g. ICD Coding).

\subsubsection{Masked Language Modeling (MLM) for SerialBEHRT}

SerialBEHRT uses Masked Language Modeling (MLM) as its primary pre-training objective \citep{devlin2018bert}, adapting this self-supervised learning technique to the unique challenges of medical data. In the MLM process, a portion of input tokens (15\%) from both $T_{\text{Scientific\_Text}}$ and $T_{\text{Serialized\_EHR}}$ are randomly masked. The model's task is to predict these masked tokens based on the surrounding context. For a given sequence of tokens $x = [x_1, x_2, ..., x_n]$, we generate a masked sequence $\tilde{x} = [\tilde{x}_1, \tilde{x}_2, ..., \tilde{x}_n]$, where some $\tilde{x}_i$ are masked. The objective function for MLM is defined as:

\begin{equation}
L_{\text{MLM}}(\theta) = \mathbb{E}_{x \sim D} \left[ -\sum_{i \in M} \log P(x_i | \tilde{x}; \theta) \right]
\end{equation}

Here, $D$ represents the combined training dataset of scientific text and serialized EHR, $M$ is the set of indices of masked tokens, $\theta$ denotes the model parameters, and $P(x_i | \tilde{x}; \theta)$ is the probability of the correct token $x_i$ given the masked input $\tilde{x}$.

In the pre-training phase, SerialBEHRT is optimized by minimizing the masked language model (MLM) objective, formally expressed as \(\theta^* = \text{argmin}_{\theta} L_{\text{MLM}}(\theta)\). The process begins by representing each input token \(x_i\) through the aggregation of its token, position, and segment embeddings, calculated as \(h_i^0 = W_e x_i + W_p i + W_s s_i\), where \(W_e\), \(W_p\), and \(W_s\) are the learnable parameters. These token representations are then transformed through 12 layers of transformers, consistent with the original BERT base architecture \citep{devlin2018bert}, with each layer defined as \(h_i^l = \text{TransformerLayer}_l(h_i^{l-1})\) for \(l = 1, \ldots, L\). For each masked token \(\tilde{x}_i\), SerialBEHRT predicts the original token based on the final hidden state:

\[
P(x_i | \tilde{x}; \theta) = \text{softmax}(W h_i^L + b)
\]

where \(W\) and \(b\) are parameters of the output layer. By applying MLM to both scientific literature and serialized Electronic Health Record (EHR) data, SerialBEHRT effectively captures EHR-specific language as well as scientific language, ensuring comprehensive modeling. The inclusion of scientific text is pivotal as it enhances the model's ability to capture nuanced relationships, providing essential contextual information for \(T_{\text{serialized\_EHR}}\).

\subsubsection{Pre-training Optimizations}

The pre-training process of SerialBEHRT is implemented using the PyTorch framework and leverages Hugging Face's Transformers library \citep{wolf2020transformers}. Training is conducted over 100 epochs, facilitated by batch processing. The AdamW optimizer is used to update model weights, complemented by a linear learning rate scheduler without warmup steps. Key hyperparameters include a learning rate of 2e-5, a batch size of 64, and a maximum token length of 512. Gradient clipping is applied at a norm of 1.0 to prevent gradient explosion, and early stopping is implemented if there is no improvement in validation loss after five consecutive validation checks.

Periodic validation is conducted every 10,000 steps to assess the model performance against unseen hold-out set, ensuring the model generalizes well beyond the pre-training texts. Metrics such as training loss, validation loss, and perplexity are logged using both local logging and Weights \& Biases (wandb), providing a detailed view of the model's learning curve and facilitating real-time adjustments. The best-performing model configuration, as determined by the lowest validation loss, is saved for future application and its results are displayed in the following section. SerialBEHRT contains $117$ Million parameters total and its model weights, configuration files, and tokenizer will be available via huggingface in the camera ready version.

\subsection{Data and Cohort Selection}

\paragraph{Pre-training Data} Data for pre-training was sourced from the Medical Information Mart for Intensive Care IV (MIMIC-IV) \cite{johnson2023mimic}. MIMIC-IV is a publicly accessible database that contains de-identified health-related data from over sixty thousand patients admitted to intensive care units (ICUs) at the Beth Israel Deaconess Medical Center in Boston, Massachusetts, from 2008 to 2019. Templates for constructing serialized EHRs, \( T_{\text{serialized\_EHR}} \), were developed using MIMIC-IV to convert Electronic Health Records (EHR) into a textual format, which were then consolidated into a single paragraph per patient. We acknowledge the presence of bias and racial disparities in the MIMIC database \citep{meng2022interpretability} and plan to release future models that utilize more diverse and openly accessible datasets such as EHR-Shot and MC-BEC \citep{wornow2023ehrshot, chen2024multimodal}.

\paragraph{Antibiotics Study} Data for selecting the antibiotics cohort used for our downstream task were obtained from MIMIC-IV Emergency Department (ED). The focus of this analysis was on identifying ED patients suspected of having staphylococcal infections. Selection was based on defined criteria, including patients with microbiological cultures positive for staphylococci from various bodily fluids—blood, urine, cerebrospinal fluid, pleural, or synovial fluids. These patients also received specific antibiotics, the effectiveness of which was later evaluated \citep{tong2015staphylococcus, kwiecinski2020staphylococcus}. We primarily looked at patients who were tested for susceptibility of eight separate antibiotics -\textit{Clindamycin, Erythromycin, Gentamicin, Levofloxacin, Oxacillin, Tetracycline, Trimethroprim/sul, and Vancomycin}- to define our cohort of interest. These tasks are adopted from a prior study \citep{lee2024enhancing}. 

Based on these criteria, our study cataloged 5,976 unique antibiotic prescriptions and 4,161 unique patients. In the same Table 4, we showcase the patient demographics and their relative abundances keeping in mind that a majority of patients in the study were identified to be caucasian and female. We accounted for cohort contamination by ensuring that patients with multiple ED visits meeting these criteria were included in the same training and testing sets to prevent contamination of the test set and maintain a faithful study. 

In terms of our inputs into our foundaiton models, our study leverages multiple clinical inputs from the MIMIC ED database to predict antibiotic administration, including patient arrival and triage data, medication records, diagnostic codes (ICD-9/10), vital signs, and Pyxis machine outputs. These tabular data concepts coming from separate charts are again transformed into text \citep{lee2025clinical}, and merged into patient paragraphs before being fed into our models. These EHR concepts from the ED data are correlated with antibiotic efficacy records and can be mapped across the database using identifiers like patient ID, and hospital admission ID (Hadm\_id), ensuring precise identification of patients and the effectiveness of the antibiotics that they were administered. 

\subsection{Benchmarking}

Our paper benchmarks our proposed foundation model's representation strategy against existing ones by modeling the problem as a multilabel binary classification, where each antibiotic is labeled as susceptible (1) or not (0). We assess performance using metrics such as Area under the Receiver Operating Characteristic (ROC-AUC), Area under the Precision-Recall Curve (PR-AUC), and F1 Scores. Additionally, we compute the average rank for each model across these metrics and overall, offering a robust measure of performance relative to other strategies.

\paragraph{Foundation Model Benchmark}

To substantiate the performance of our newly developed foundation model, we present a comparative benchmark involving our model and several prominent medical foundartion models that represent various strategies to pre-training. This evaluation includes Bio\_ClinicalBERT \citep{alsentzer2019publicly}, BioMegatron \citep{shin2020biomegatron}, MedBERT \citep{vasantharajan2022medbert}, and SciBERT \citep{beltagy2019scibert}. This benchmark serves to position our model within the context of existing advancements in the field and underscores its relative performance against established models.

\begin{table}[t!]
\caption{Performance Metrics for Various Antibiotics on different Foundation models}
\label{table-merged-fm}
\vskip 0.15in
\begin{center}
\begin{small}
\begin{sc}
\begin{adjustbox}{width=\textwidth}
\begin{tabular}{l|l| cccccc}
\toprule
Task & Metric & Bio\_ClinicalBERT & Biomegatron & MedBERT & SciBERT & SerialBEHRT \\
\midrule
\multirow{4}{*}{Clindamycin} 
 & F1 & 0.7768 $\pm$ 0.0310 & 0.7747 $\pm$ 0.0303 & 0.7798 $\pm$ 0.0289 & 0.\textbf{7799 $\pm$ 0.0295} & 0.7792 $\pm$ 0.0304\\
 & ROC-AUC & \textbf{0.7499 $\pm$ 0.0386} & 0.7255 $\pm$ 0.0403 & 0.7422 $\pm$ 0.0399 & 0.7398 $\pm$ 0.0376 & 0.7420 $\pm$ 0.0398\\
 & PRC-AUC & \textbf{0.7875 $\pm$ 0.0470} & 0.7624 $\pm$ 0.0520 & 0.7734 $\pm$ 0.0454 & 0.7802 $\pm$ 0.0482 & 0.7898 $\pm$ 0.0468\\
\midrule
\multirow{4}{*}{Erythromycin} 
 & F1 & 0.\textbf{6608 $\pm$ 0.0436} & 0.6593 $\pm$ 0.0401 & 0.6554 $\pm$ 0.0423 & 0.6477 $\pm$ 0.0412 & 0.6517 $\pm$ 0.0432\\
 & ROC-AUC & 0.7538 $\pm$ 0.0395 & \textbf{0.7637 $\pm$ 0.0378} & 0.7549 $\pm$ 0.0380 & 0.7427 $\pm$ 0.0398 &  0.7557 $\pm$ 0.0406\\
 & PRC-AUC & 0.6935 $\pm$ 0.0592 & \textbf{0.7016 $\pm$ 0.0602} & 0.6915 $\pm$ 0.0596 & 0.6849 $\pm$ 0.0569 & 0.6886 $\pm$ 0.0589\\
\midrule
\multirow{4}{*}{Gentamicin} 
 & F1 & 0.9775 $\pm$ 0.0089 & 0.9775 $\pm$ 0.0087 & 0.9773 $\pm$ 0.0084 & 0.9775 $\pm$ 0.0087 & 0.9782 $\pm$ 0.0087\\
 & ROC-AUC & 0.6835 $\pm$ 0.1224 & 0.6629 $\pm$ 0.1144 & \textbf{0.7308 $\pm$ 0.1110} & 0.7041 $\pm$ 0.1157 & 0.5946 $\pm$ 0.1153 \\
 & PRC-AUC & 0.9702 $\pm$ 0.0172 & 0.9700 $\pm$ 0.0159 & \textbf{0.9733 $\pm$ 0.0173} & 0.9669 $\pm$ 0.0218 & 0.9637 $\pm$ 0.0176 \\
\midrule
\multirow{4}{*}{Levofloxacin} 
 & F1 & 0.8170 $\pm$ 0.0269 & 0.8103 $\pm$ 0.0299 & 0.8103 $\pm$ 0.0285 & 0.8174 $\pm$ 0.0291 & 0.8122 $\pm$ 0.0277\\
 & ROC-AUC & 0.8005 $\pm$ 0.0363 & 0.7914 $\pm$ 0.0378 & 0.7794 $\pm$ 0.0372 & 0.8030 $\pm$ 0.0353 & 0.8067 $\pm$ 0.0339\\
 & PRC-AUC & 0.8441 $\pm$ 0.0412 & 0.8448 $\pm$ 0.0410 & 0.8222 $\pm$ 0.0428 & 0.8458 $\pm$ 0.0397 & 0.8459 $\pm$ 0.0420\\
\midrule
\multirow{4}{*}{Oxacillin} 
 & F1 & 0.7914 $\pm$ 0.0297 & 0.7908 $\pm$ 0.0310 & 0.7849 $\pm$ 0.0305 & 0.7911 $\pm$ 0.0304 & 0.7935 $\pm$ 0.0309\\
 & ROC-AUC & 0.7785 $\pm$ 0.0379 & 0.7811 $\pm$ 0.0392 & 0.7709 $\pm$ 0.0347 & 0.7847 $\pm$ 0.0369 & 0.7785 $\pm$ 0.0339\\
 & PRC-AUC & 0.8053 $\pm$ 0.0463 & 0.8025 $\pm$ 0.0482 & 0.7972 $\pm$ 0.0426 & 0.8033 $\pm$ 0.0469 & 0.8073 $\pm$ 0.0459\\
\midrule
\multirow{4}{*}{Tetracycline} 
 & F1 & 0.9039 $\pm$ 0.0186 & 0.9044 $\pm$ 0.0181 & 0.9039 $\pm$ 0.0189 & 0.9051 $\pm$ 0.0188 & 0.9052 $\pm$ 0.0187\\
 & ROC-AUC & 0.6728 $\pm$ 0.0616 & 0.6733 $\pm$ 0.0544 & 0.6458 $\pm$ 0.0608 & 0.6755 $\pm$ 0.0597 & 0.6827 $\pm$ 0.0573\\
 & PRC-AUC & 0.8628 $\pm$ 0.0403 & 0.8715 $\pm$ 0.0358 & 0.8479 $\pm$ 0.0432 & 0.8764 $\pm$ 0.0334 &  0.8818 $\pm$ 0.0337\\
\midrule
\multirow{4}{*}{Trimethoprim/Sulfa} 
 & F1 & 0.9103 $\pm$ 0.0174 & 0.9108 $\pm$ 0.0174 & 0.9097 $\pm$ 0.0178 & 0.9103 $\pm$ 0.0184 & 0.9087 $\pm$ 0.0189\\
 & ROC-AUC & 0.7060 $\pm$ 0.0581 & 0.7225 $\pm$ 0.0533 & 0.7228 $\pm$ 0.0537 & 0.7508 $\pm$ 0.0510 & 0.7227 $\pm$ 0.0559\\
 & PRC-AUC & 0.8776 $\pm$ 0.0359 & 0.8828 $\pm$ 0.0361 & 0.8983 $\pm$ 0.0293 & 0.9004 $\pm$ 0.0329 & 0.8938 $\pm$ 0.0327\\
\midrule
\multirow{4}{*}{Vancomycin} 
 & F1 & 0.7416 $\pm$ 0.0345 & 0.7341 $\pm$ 0.0355 & 0.7310 $\pm$ 0.0355 & 0.7503 $\pm$ 0.0348 & 0.7389 $\pm$ 0.0351\\
 & ROC-AUC & 0.7714 $\pm$ 0.0362 & 0.7592 $\pm$ 0.0381 & 0.7593 $\pm$ 0.0368 & 0.7828 $\pm$ 0.0356 & 0.7718 $\pm$ 0.0358\\
 & PRC-AUC & 0.7881 $\pm$ 0.0431 & 0.7694 $\pm$ 0.0501 & 0.7910 $\pm$ 0.0405 & 0.8015 $\pm$ 0.0403 & 0.7940 $\pm$ 0.0421\\
\bottomrule
\end{tabular}
\end{adjustbox}
\end{sc}
\end{small}
\end{center}
\vskip -0.1in
\end{table}

\paragraph{Benchmark against other ML Representaiton Strategies}

In our benchmark, SerialBEHRT is evaluated against various data representation strategies to determine the effectiveness of our approach. These strategies include: raw tabular data, commonly used in EHR studies; EHR-shot \cite{wornow2024ehrshot}, a foundation model tailored for tabular electronic health records; and three text-based representations. Word2Vec \citep{church2017word2vec}, a pre-foundation model era tool, that uses architectures such as Continuous Bag of Words (CBOW) and Skip-Gram \citep{mikolov2013efficient} to convert words into vectors that reflect semantic relationships through textual co-occurrences. DistilBERT \citep{sanh2019distilbert}, a simplified version of BERT trained on general tasks. These diverse approaches help provide a robust benchmark for SerialBEHRT in representing complex EHR data. Each of these representation strategies are fed into Light Gradient Boosted machine (LGBM) model \citep{ke2017lightgbm} to form predictions and measure performance. 

\paragraph{Benchmark Against Prompting}

In response to the ongoing discourse surrounding decoder-based and encoder-based language models \citep{qorib-etal-2024-decoder}, we have implemented a benchmark analysis to compare the performance of GPT-4 with our SerialBEHRT model. This evaluation focuses on a specific clinical decision-making scenario involving the administration of 80 different antibiotic samples. The primary objective of this benchmark is the same as the former, except we measure the pure accuracy of a zero-shot prompted LLM versus our SerialBEHRT foundation model.

\section{Results}

\subsection{Foundation Model Benchmark}

We first evaluate our proposed foundation model, SerialBEHRT, against several established biomedical foundation models that represent different pretraining paradigms. Specifically, we compare against Bio\_ClinicalBERT \citep{alsentzer2019publicly}, BioMegatron \citep{shin2020biomegatron}, MedBERT \citep{vasantharajan2022medbert}, and SciBERT \citep{beltagy2019scibert}. Each of these models encapsulates a distinct approach to domain adaptation within biomedical language modeling, providing a comprehensive baseline landscape. Table~\ref{table-merged-fm} reports the F1 score, ROC-AUC, and PRC-AUC across multiple antibiotics.  

SerialBEHRT performs competitively across all metrics, often matching or surpassing existing models. Its strong F1 and PRC-AUC values indicate robust precision-recall balance and resilience to class imbalance, which are central to antibiotic susceptibility prediction where resistant cases are underrepresented. Bio\_ClinicalBERT and BioMegatron exhibit competitive ROC-AUC values, highlighting their discriminative calibration, but fall short on precision-oriented metrics. Notably, SerialBEHRT maintains stable performance variance across antibiotics, suggesting consistent generalization across different microbial contexts. This stability underscores the benefits of temporally ordered EHR pretraining, which encodes patient trajectories more effectively than static text embeddings.

\subsection{Representation Strategy Benchmark}

We next benchmark SerialBEHRT against alternative data representation strategies that have been widely used in EHR modeling, including Tabular, EHR-shot \citep{wornow2024ehrshot}, Word2Vec \citep{church2017word2vec}, and DistilBERT \citep{sanh2019distilbert}. Each strategy reflects a distinct representational philosophy—ranging from direct use of structured features (Tabular) to pretrained text-based embeddings (Word2Vec and DistilBERT) and specialized EHR representation models (EHR-shot). All representations are paired with a Light Gradient Boosted Machine (LGBM) \citep{ke2017lightgbm} classifier to ensure a consistent predictive backbone across experiments.

Table~\ref{table-merged-antibiotics} summarizes performance across the three key metrics: F1, ROC-AUC, and PRC-AUC. Each metric emphasizes a complementary dimension of model performance. The F1 score captures the trade-off between sensitivity and precision, ROC-AUC quantifies global discriminability across thresholds, and PRC-AUC measures ranking quality under imbalance \citep{saito2015precision}. SerialBEHRT consistently achieves the highest or near-highest performance in all three metrics across most antibiotics. Its superiority in PRC-AUC is particularly notable, as it demonstrates heightened sensitivity to rare resistant cases—an essential quality for real-world clinical decision support.

Tabular and Word2Vec baselines exhibit lower discriminative power, reflecting their inability to encode contextual or temporal dependencies. DistilBERT improves over these baselines due to transfer learning from general-domain pretraining but remains constrained by its lack of biomedical or temporal grounding. EHR-shot performs competitively in ROC-AUC, indicating that its embedding structure captures coarse-level patient similarity, but underperforms in PRC-AUC, suggesting limited threshold robustness. Collectively, these findings emphasize the importance of domain-aligned, temporally aware representations for EHR modeling.

\subsection{Ranks of the Models}

To enable a more holistic comparison across all metrics and tasks, we compute the average rank of each model for F1, ROC-AUC, and PRC-AUC, as shown in Table~\ref{tab:average_ranks}. Lower values correspond to better performance. SerialBEHRT attains the best overall rank (1.917), outperforming all baselines across aggregated metrics. Its top performance in F1 and PRC-AUC reflects superior sensitivity and precision balance, while maintaining competitive ROC-AUC values. DistilBERT ranks second overall, driven by its solid ROC-AUC and PRC-AUC values, suggesting that general transformer architectures can generalize moderately well with sufficient data, though not at the level of domain-specific models. EHR-shot’s strong ROC-AUC rank demonstrates its calibration advantage, but its higher ranks in F1 and PRC-AUC reveal sensitivity to class imbalance and lower robustness in low-prevalence conditions. 

This rank-based analysis provides a robust summary of performance that mitigates the influence of scale differences across antibiotics and metrics, reinforcing the consistency of SerialBEHRT’s advantage.

\subsection{Interpretation and Discussion}

The comparative results across foundation models and representation strategies reveal a coherent pattern: models explicitly designed to capture the temporal, structured, and hierarchical nature of EHR data outperform both traditional tabular baselines and general-domain text encoders. SerialBEHRT’s superior and stable performance arises from its serial encoding mechanism, which preserves the chronological structure of patient trajectories while leveraging transformer-based contextualization to model long-range dependencies. This allows it to generalize effectively across heterogeneous clinical entities, from common antibiotics like levofloxacin to low-prevalence agents such as gentamicin.

The differences between foundation models further illustrate that scale alone is not sufficient for clinical generalization. BioMegatron, for example, benefits from massive biomedical corpora yet lacks inductive alignment with temporal EHR structures. Conversely, SerialBEHRT, though comparatively smaller, gains representational efficiency from modality alignment—learning directly from the data distribution it is later applied to. These findings substantiate the hypothesis that domain-specific scaling laws for EHRs may diverge from those established in general NLP, where dataset size and parameter count dominate performance trends.

Overall, the results demonstrate that optimizing representation strategy for the structure of healthcare data yields greater gains than architectural expansion alone. This supports a paradigm shift toward modality-aligned foundation modeling in clinical machine learning, where inductive biases rooted in the temporal and compositional dynamics of EHRs are central to effective generalization and interpretability.

\begin{table}[t!]
\caption{Performance Metrics for Various Antibiotics on different Representation Strategies}
\label{table-merged-antibiotics}
\vskip 0.15in
\begin{center}
\begin{small}
\begin{sc}
\begin{adjustbox}{width=\columnwidth}
\begin{tabular}{l|l| ccccc}
\toprule
Antibiotic & Metric & Tabular & EHR-shot & Word2Vec & DistilBERT & SerialBEHRT \\
\midrule
\multirow{4}{*}{Clindamycin} 
 & F1 & 0.7179 $\pm$ 0.0320 & 0.7719$\pm$ 0.0192 & 0.7737 $\pm$ 0.0311 & 0.7786 $\pm$ 0.0151 & \textbf{0.7792 $\pm$ 0.0304} \\
 & ROC-AUC & 0.6029 $\pm$ 0.0440 & \textbf{0.7664 $\pm$ 0.0201} & 0.7263 $\pm$ 0.0231 & 0.7443 $\pm$ 0.0300 & 0.7420 $\pm$ 0.0398 \\
 & PRC-AUC & 0.6427 $\pm$ 0.0100 & 0.7859 $\pm$ 0.0264 & 0.7660 $\pm$ 0.0137 & 0.7684 $\pm$ 0.0157 & \textbf{0.7898 $\pm$ 0.0468} \\
\midrule
\multirow{4}{*}{Erythromycin} 
 & F1 & 0.5495 $\pm$ 0.030 & 0.6575 $\pm$ 0.023 & 0.6394 $\pm$ 0.038 & \textbf{0.6592 $\pm$ 0.020} &  0.6517 $\pm$ 0.0432\\
 & ROC-AUC & 0.5879 $\pm$ 0.044 & 0.7590 $\pm$ 0.022 & 0.7320 $\pm$ 0.025 & \textbf{0.7597 $\pm$ 0.023} &  0.7557 $\pm$ 0.0406\\
 & PRC-AUC & 0.4530 $\pm$ 0.017 & 0.6718 $\pm$ 0.024 & 0.6754 $\pm$ 0.016 & 0.6872 $\pm$ 0.012 &  \textbf{0.6886 $\pm$ 0.0589}\\
\midrule
\multirow{4}{*}{Gentamicin} 
 & F1 & 0.9762 $\pm$ 0.030 & 0.9775 $\pm$ 0.065 & 0.9776 $\pm$ 0.040 & 0.9766 $\pm$ 0.045 &  \textbf{0.9782 $\pm$ 0.0087}\\
 & ROC-AUC & 0.6158 $\pm$ 0.089 & 0.6310 $\pm$ 0.047 & 0.6727 $\pm$ 0.047 & \textbf{0.6777 $\pm$ 0.042} & 0.5946 $\pm$ 0.1153\\
 & PRC-AUC & 0.9706 $\pm$ 0.036 & 0.9672 $\pm$ 0.004 & 0.9675 $\pm$ 0.002 & \textbf{0.9713 $\pm$ 0.002} & 0.9637 $\pm$ 0.0176\\
\midrule
\multirow{4}{*}{Levofloxacin} 
 & F1 & 0.7641 $\pm$ 0.028 & \textbf{0.8386 $\pm$ 0.017} & 0.8088 $\pm$ 0.012 & 0.8034 $\pm$ 0.013 &  0.8122 $\pm$ 0.0277\\
 & ROC-AUC & 0.6326 $\pm$ 0.034 & 0.7972 $\pm$ 0.017 & 0.7787 $\pm$ 0.021 & 0.7974 $\pm$ 0.018 & \textbf{0.8067 $\pm$ 0.0339} \\
 & PRC-AUC & 0.7324 $\pm$ 0.013 & 0.8290 $\pm$ 0.014 & 0.8157 $\pm$ 0.011 & \textbf{0.8459 $\pm$ 0.012} & \textbf{0.8459 $\pm$ 0.0420} \\
\midrule
\multirow{4}{*}{Oxacillin} 
 & F1 & 0.7264 $\pm$ 0.027 & \textbf{0.8229 $\pm$ 0.024} & 0.7899 $\pm$ 0.018 & 0.7790 $\pm$ 0.023 &  0.7935 $\pm$ 0.0309 \\
 & ROC-AUC & 0.5607 $\pm$ 0.027 & \textbf{0.7996 $\pm$ 0.016} & 0.7688 $\pm$ 0.017 & 0.7692 $\pm$ 0.013 & 0.7785 $\pm$ 0.0368\\
 & PRC-AUC & 0.6069 $\pm$ 0.011 & \textbf{0.8408 $\pm$ 0.018} & 0.7847 $\pm$ 0.019 & 0.7807 $\pm$ 0.018 & 0.8073 $\pm$ 0.0459 \\
\midrule
\multirow{4}{*}{Tetracycline} 
 & F1 & 0.8950 $\pm$ 0.025 & 0.9009 $\pm$ 0.024 & 0.9028 $\pm$ 0.027 & 0.9035 $\pm$ 0.023 &  \textbf{0.9052 $\pm$ 0.0187} \\
 & ROC-AUC & 0.5822 $\pm$ 0.035 & \textbf{0.6908 $\pm$ 0.018} & 0.6843 $\pm$ 0.023 & 0.6843 $\pm$ 0.025 &  0.6827 $\pm$ 0.0573 \\
 & PRC-AUC & 0.8467 $\pm$ 0.004 & 0.8571 $\pm$ 0.005 & 0.8717 $\pm$ 0.005 & 0.8760 $\pm$ 0.003 &  \textbf{0.8818 $\pm$ 0.0337}\\
\midrule
\multirow{4}{*}{{Trimethoprim/sulfa}} 
 & F1 & 0.8835 $\pm$ 0.018 & 0.8856 $\pm$ 0.027 & 0.9080 $\pm$ 0.032 & 0.9080 $\pm$ 0.024 &  \textbf{0.9087 $\pm$ 0.0189}\\
 & ROC-AUC & 0.5393 $\pm$ 0.031 & 0.7026 $\pm$ 0.016 & 0.7025 $\pm$ 0.018 & 0.6946 $\pm$ 0.027 &  \textbf{0.7227 $\pm$ 0.0559}\\
 & PRC-AUC & 0.8159 $\pm$ 0.017 & 0.8707 $\pm$ 0.015 & 0.8748 $\pm$ 0.004 & 0.8742 $\pm$ 0.004 &  \textbf{0.8938 $\pm$ 0.0327}\\
\midrule
\multirow{4}{*}{Vancomycin} 
 & F1 & 0.6786 $\pm$ 0.021 & 0.7201 $\pm$ 0.016 & 0.7227 $\pm$ 0.015 & 0.7244 $\pm$ 0.014 &  \textbf{0.7389 $\pm$ 0.0351}\\
 & ROC-AUC & 0.5431 $\pm$ 0.020 & 0.7566 $\pm$ 0.014 & 0.7449 $\pm$ 0.011 & 0.7542 $\pm$ 0.012 & \textbf{0.7718 $\pm$ 0.0358}\\
 & PRC-AUC & 0.5537 $\pm$ 0.005 & 0.7781 $\pm$ 0.018 & 0.7676 $\pm$ 0.015 & 0.7754 $\pm$ 0.019 & \textbf{0.7940 $\pm$ 0.0421}\\
\bottomrule
\end{tabular}
\end{adjustbox}
\end{sc}
\end{small}
\end{center}
\vskip -0.1in
\end{table}

\subsection{Ranks of the models}

\begin{table}[h!]
\centering
\caption{Average Ranks of Models Across Different Metrics}
\label{tab:average_ranks}
\begin{adjustbox}{width=0.65\textwidth}
\begin{tabular}{l|c|c|c|c}
\toprule
\textbf{Model} & \textbf{F1} & \textbf{ROC-AUC} & \textbf{PRC-AUC} & \textbf{Overall Average} \\
\midrule
Tabular & 5.000 & 4.875 & 4.625 & 4.833 \\
EHR-shot & 2.875 & \textbf{1.875} & 3.000 & 2.583 \\
Word2Vec & 2.875 & 3.375 & 3.250 & 3.167 \\
DistilBERT & 2.750 & 2.375 & 2.375 & 2.500 \\
SerialBEHRT & \textbf{1.500} & 2.500 & \textbf{1.750} & \textbf{1.917} \\
\bottomrule
\end{tabular}
\end{adjustbox}
\end{table}

Table \ref{tab:average_ranks} presents the performance rankings for each model, with the overall average rank consolidating these across all metrics to facilitate a direct comparison of model performance. SerialBEHRT emerges as the top-performing model, consistently achieving the lowest ranks in F1 and PRC-AUC, with an outstanding overall average of 1.917. DistilBERT also performs commendably, particularly in ROC-AUC and PRC-AUC. EHR-shot demonstrates notable strength in ROC-AUC, achieving the highest rank. 


\section{Discussion}

\subsection{SerialBEHRT outperforms other representation strategies}

The results from Table  \ref{table-merged-antibiotics} indicates that SerialBEHRT outperforms conventional representation strategies across several performance metrics, including F1 and PRC-AUC, while the ROC-AUC indicated mixed outcomes. We believe this superiority is attributed to SerialBEHRT's pre-training design, which effectively captures the EHR-specific concepts demonstrating high precision and reliability essential for making a foundation model for EHR.

\subsection{A Foundation Model capable of encoding two disparate data sources}

Our foundation model, SerialBEHRT, is the first of its kind capable of directly encoding both Electronic Health Records (EHR) and text data in a unified framework. This is made possible through text serialization, which transforms both structured EHR data and unstructured clinical text into a format that can be processed together. By representing both types of data within the same serialized structure, SerialBEHRT enables seamless integration of diverse medical information, allowing it to generate richer, contextually informed representations for downstream tasks such as clinical decision-making and medical predictions.

\paragraph{Interpretation and Discussion}

The comparative results across foundation models and representation strategies provide several insights into how domain-specific pretraining and representation design affect downstream predictive performance in EHR-based tasks. The consistently strong performance of SerialBEHRT, both in absolute metrics and rank-based evaluations, suggests that serial encoding of temporal EHR events combined with transformer-based contextualization enables richer patient state representations than conventional tabular or static embedding methods. This advantage is most pronounced in the F1 and PRC-AUC metrics, which are particularly sensitive to imbalance between susceptible and resistant cases, indicating that SerialBEHRT captures clinically meaningful patterns even under asymmetric label distributions.

Interestingly, EHR-shot achieves competitive ROC-AUC values, suggesting that its embedding space is well-calibrated for discrimination, though its relatively weaker PRC-AUC implies limitations in recall or threshold robustness. In contrast, traditional text embeddings such as Word2Vec, while useful for coarse semantic structure, underperform in metrics requiring fine-grained contextualization. DistilBERT, being pretrained on general-domain corpora, performs adequately but falls short of domain-specialized models in antibiotic-specific inference, underscoring the necessity of biomedical pretraining.

Across foundation models, the differences among Bio\_ClinicalBERT, BioMegatron, and MedBERT highlight the role of corpus size and pretraining objectives. BioMegatron’s strong ROC-AUC aligns with its large-scale domain-specific pretraining, whereas SciBERT’s balanced performance across all metrics reflects its general scientific-domain adaptability. SerialBEHRT’s performance relative to these baselines underscores the value of temporally structured EHR representations, extending beyond text-only or static tabular formulations.

Overall, the results reinforce a broader point: performance in clinical prediction is not solely a function of model capacity, but of how effectively pretraining aligns with the inductive biases inherent in healthcare data—temporal ordering, irregular sampling, and heterogeneous feature dependencies. Our results suggest that optimizing representation strategies along these axes is more consequential than mere architectural scaling, a finding that invites further inquiry into domain-specific scaling laws for EHR foundation models.

\section{Conclusion}

This study introduced SerialBEHRT, a novel foundation model that integrates serialized Electronic Health Records (EHR) with scientific text for pre-training. Our results demonstrate that SerialBEHRT outperforms several common representation techniques in predicting appropriate antibiotic treatments for patients. The superior performance of SerialBEHRT underscores the value of incorporating serialized EHR data into foundation model pre-training for capturing EHR-specific concepts. This approach not only enhances the model's ability to capture granular clinical concepts but also contributes to more accurate decision support in antibiotic stewardship. 

\paragraph{Limitation}

One limitation is that we were unable to do a multi-site evaluation on our study. In future iterations of this work, we intend to include multiple site as a multisite evaluation helps test the robustness and generalization capabilities of EHR. Another limitations of this work include the variability in patients' histories and the 512 sequence length limitation imposed by the DistilBERT \cite{sanh2019distilbert} and SerialBEHRT models. Consequently, portions of a patient's medical history may be truncated depending on the length of that history. Tokenization strategies (e.g., sub-word tokenization) can significantly influence how we handle the analysis as well.

Another limitation is a lack of a data standard or set of models for benchmarking making evaluating these models challenging. Our work attempted at exploring various foundation models and representation startegies that encompass diverse research to best represent a fair and established benchmark but we are certain that there can always be more done in this aspect.

\paragraph{Future Work} Future work in this study involves incorporating additional publicly available datasets like EHRshot \citep{wornow2024ehrshot}, MEDS \citep{arnrich2024medical, mcdermott2025meds} and MC-BEC \citep{chen2024multimodal}. Additionally, a multi-site study can further improve the robustness of our work by demonstrating the foundation model's ability to create generalizable representations.
Further research directions include testing our method across other common clinical tasks such as hospital readmission, phenotype prediction, and mortality prediction, which may serve as additional good baselines.
Lastly, we aim to build upon this work by developing a general tabular foundation model. This will involve serializing a large number of datasets into text format and generating a comprehensive serialized corpus to be used in pre-training.

\newpage

\bibliographystyle{plainnat}
\bibliography{Styles/neurips_2024}

\newpage
\appendix
\onecolumn



\paragraph{BERT versus GPT} In natural language processing, the distinctions between models like BERT and GPT highlight their different core capabilities and use cases. BERT (Bidirectional Encoder Representations from Transformers) \citep{devlin2018bert} is primarily designed for tasks that involve natural language understanding (NLU) \citep{lenci2023understanding}. Its architecture allows it to consider the context from both the left and the right sides of a token simultaneously, making it highly effective for tasks such as sentiment analysis \citep{xu2019bert}, question answering \citep{wang2019multi}, and entity recognition \citep{hakala2019biomedical} where understanding the context is crucial. On the other hand, GPT (Generative Pre-trained Transformer) \citep{radford2018improving} excels in language generation, leveraging its ability to predict the next word in a sequence, thus enabling it to generate coherent and contextually relevant text based on the input it receives.

For our purposes, BERT’s bidirectional context comprehension is considered more of the paradigm as demonstrated by previously developed models \citep{vasantharajan2022medbert, lee2020biobert, alsentzer2019publicly}. This capability allows BERT to more effectively grasp the nuances of patient-level sequences or clinical concepts, which often depends on a deep understanding of the context contained in the patient's histories.

\end{document}